\documentclass[runningheads]{llncs}
\usepackage{graphicx}
\usepackage[linesnumbered,ruled]{algorithm2e}
\usepackage{amsmath}
\usepackage{amssymb}
\usepackage{booktabs}
\usepackage{color}
\pdfoutput=1

\newsavebox{\algleft}
\newsavebox{\algright}

\newcommand{\calD}{\ensuremath{\mathcal{D}}}
\newcommand{\calF}{\ensuremath{\mathcal{F}}}

\newcommand{\calJ}{\ensuremath{\mathcal{J}}}
\newcommand{\calL}{\ensuremath{\mathcal{L}}}
\newcommand{\calM}{\ensuremath{\mathcal{M}}}
\newcommand{\calR}{\ensuremath{\mathcal{R}}}

\newcommand{\R}{\ensuremath{\mathbb{R}}}
\newcommand{\F}{\ensuremath{\calF}}
\newcommand{\M}{\ensuremath{\calM}}
\newcommand{\y}{\ensuremath{y}}

\definecolor{col1}{RGB}{228,32,50}
\definecolor{col2}{RGB}{236,116,4}
\definecolor{col3}{RGB}{250,187,0}
\definecolor{col4}{RGB}{59,178,160}
\definecolor{col5}{RGB}{0,106,163}
\definecolor{col_mean}{RGB}{0,70,114}

\begin{document}
\title{mlVIRNET: Multilevel Variational Image Registration Network}
%
%
\author{Alessa Hering\inst{1,2}, Bram van Ginneken\inst{2} \and
Stefan Heldmann\inst{1}}

\authorrunning{A. Hering et al.}
%
\institute{Fraunhofer MEVIS, L{\"u}beck, Germany
\email{alessa.hering@mevis.fraunhofer.de}\\
\and
Diagnostic Image Analyse Group, Radboudumc, Nijmegen, Netherlands\\
}
\maketitle              
\begin{abstract}
We present a novel multilevel approach for deep learning based image registration. Recently published deep learning based registration methods have shown promising results for a wide range of tasks. However, these algorithms are still limited to relatively small deformations. Our method addresses this shortcoming by introducing a multilevel framework, which computes deformation fields on different scales, similar to conventional methods. Thereby, a coarse-level alignment is obtained first, which is subsequently improved on finer levels. We demonstrate our method on the complex task of inhale-to-exhale lung registration. We show that the use of a deep learning multilevel approach leads to significantly better registration results.

\end{abstract}
\section{Introduction}
Image registration is the process of aligning two or more images to achieve point-wise spatial correspondence. This is a fundamental step for many medical image analysis tasks and has been an active field of research for decades. Since recently, deep learning based approaches have been successfully employed for image registration \cite{balakrishnan2019voxelmorph,eppenhof2019progressively,hering2019unsupervised,HeringEtAl2019,rohe2017svf,vos2019DLIR}. They have shown promising results in a wide range of application. However, capturing large motion and deformation with deep learning based registration is still an open challenge. In common iterative image registration approaches, this is typically addressed with a multilevel coarse-to-fine registration strategy \cite{bajcsy1989multiresolution,haber2004cofir,schnabel2001generic}. Starting on a coarse grid with smoothed and down-sampled versions of the input images a deformation field is computed which is subsequently prolongated on the next finer level as a initial guess. Hereby, a coarse level alignment is obtained first that typically captures the large motion components and which is later improved on finer levels for the alignment of more local details. Most of the recently presented deep learning based approaches also make use of a multilevel strategy as they are based on the U-Net architecture \cite{balakrishnan2019voxelmorph,hering2019unsupervised,HeringEtAl2019,rohe2017svf}. Thereby, the first half of the "U" is used to generate features on different scales starting at the highest resolution and reducing the resolution through pooling operations. In this procedure, however, only feature maps on different levels are calculated but neither different image resolutions are used nor deformation fields are computed. Only a few approaches  implement a multi-resolution or hierarchical strategy in the sense of multilevel strategies associated with  conventional methods. In \cite{hu2018labeldriven} the authors proposed an architecture which is divided into a global and a local network, which are optimized together. In \cite{eppenhof2019progressively} a multilevel strategy is incorporated into the training of a  U-net. Here, a CNN is grown and trained progressively level-by-level. In \cite{vos2019DLIR} a patch based  approach is presented, where multiple  CNNs (ConvNets) are combined additive into a larger architecture for performing coarse-to-fine image registration of patches. The results from the  patches are then combined into a deformation field warping the whole image. In this work, we address this challenge and present a multilevel strategy for deep learning based image registration to advance state-of-the-art approaches.
The contribution of this paper includes:
\begin{itemize}
\item We present deep learning based multilevel registration that is able to compensate and handle large deformations by computing deformation fields on different scales and functionally compose them. 

\item Our method is a theoretically sound and a direct transference of coarse-to-fine registration from conventional, iterative registration schemes to the deep learning based methods.

\item We do not rely on patches. We take the whole image information into account and always consider the full field of view on all levels. 

\item A robust and fast registration method for the complex task of inhale-to-exhale registration validated on a large dataset of 270  thoracic CT scan pairs of the multi-center COPDGene study and on the publicly available DIR-LAB dataset \cite{castillo2013DIRLAB}.
\end{itemize}
\section{Method}
Our deep learning based framework for deformable image registration consists of two main building blocks. The first one is the specific design of the convolutional neural network and the loss function. In general, several architectures together with different distance measures, regularizer and penalty terms can be used. However, we focus on a U-Net based architecture, combined with a loss function that has shown good results for the task of pulmonary registration \cite{ruhaak2017estimation}. The second main building block is the embedding into a multilevel approach from coarse to fine. In the following, we give a brief outline of the variational setup, then we describe our particular architecture and loss function and, finally, we present its embedding into a multilevel approach.  

\subsection*{Variational Registration Approach:} 
Following \cite{Modersitzki2009}, let $\calF,\calM:\R^3\to\R$ denote the fixed image and moving image, respectively, and let $\Omega\subset\R^3$ be a domain modeling the field of view of $\calF$. We aim to compute a deformation $\y:\Omega\to\R^3$ that aligns the fixed image $\calF$ and the moving image $\calM$ on the field of view $\Omega$ such that $\calF(x)$ and $\calM(\y(x))$ are similar for $x\in\Omega$. The deformation is defined as a mimimizer of a suitable cost function that typically takes the form
\begin{equation}\label{eq:Jfun}
\calJ(\calF,\calM,\y)~=~\calD(\calF,\calM(\y))~+~\alpha\calR(\y)
\end{equation} 
with so-called distance measure $\calD$ that quantifies the similarity of fixed image $\calF$ and deformed moving image $\calM(y)$ and so-called a regularizer $\calR$ that forces smoothness of the deformation typically by penalizing of spatial derivatives. Typical examples for the distance measure are, e.g., the squared $L_2$ norm of the difference image (SSD), cross correlation (CC) or mutual information (MI). In our experiments, we follow the approach of \cite{ruhaak2017estimation} using the edge based normalized gradient fields distance measure (NGF) and second order curvature regularization.

\subsection*{Deep Learning based Image Registration} 
In contrast to conventional registration \cite{Modersitzki2009}, we do not employ iterative optimization during inference of new unseen images but use a convolutional neural network (CNN) that takes images $\F$ and $\M$ as input and yields the deformation $\y$ as output. Thus, in the context of CNNs we can consider $\y$ as a function of a trainable CNN model parameter vector $\theta\in\R^{P}$ and input images \F, \M, i.e. $ \y(x) \equiv \y(\theta;\F,\M,x)$. In an unsupervised learning approach, we set up a loss function $\calL$ that depends on $\calF$, $\calM$ and $\y$, and then $\theta$  is learned by training, i.e., minimizing the expected value of $\calL$ among a set of representative input images w.r.t. $\theta$. A natural choice would $\calL=\calJ$. However, in our particular application, we have additional information available during training and we perform a weakly supervised approach. To this end, we define our loss function as suggested in \cite{HeringEtAl2019}
\begin{equation}
\calL(\calF,\calM,b_\calF,b_\calM,\y) = \calJ(\calF,\calM,\y) + \frac\beta2\|b_\calF - b_\calM(\y)\|_{L_2}^2
\end{equation} 
with binary segmentation masks $b_\calF$ and $b_\calM(y)$ of the fixed and warped moving image, respectively. Note that these segmentations are only used to evaluate the loss function for training and their are not used as network input. 

\subsection*{Single Level Architecture}
Our CNN $\y\equiv \y(\theta,\M,\F)$ is based on a U-Net which takes the concatenated 3D moving and fixed image as input and predicts a 3D dense displacement field. The network consists of three resolution levels starting with 16 filters in the first layer, which are doubled after each downsampling step. We apply 3D convolutions in both encoder and decoder stage with a kernel size of 3 followed by a batch normalization and a ReLU layer. For downsampling the feature maps during the encoder path, an $2\times 2\times 2$ average pooling operation with a stride of 2 is used. Transposed convolutions upsample and halve the feature maps in the decoder path. At the final layer, a $1\times 1\times 1$ convolution is used to map each 16 component feature vector to a three dimensional displacement vector. 

\subsection*{Multilevel Deep Learning based Registration}
Multilevel continuation and  scale space techniques have been proven very efficient in conventional variational registration approaches to avoid local minima, to reduce topological changes or foldings and to speed up runtimes \cite{bajcsy1989multiresolution,kabus2010multilevel,haber2004cofir,schnabel2001generic}. 
However, beside carrying over these properties, our major motivation here is, to overcome the limitation of deep learning based registration to small and local deformations. 

 We follow the ideas of standard multilevel registration and compute coarse grid solutions that are prolongated and refined on the subsequent finer level. To this end, first we create image pyramids $\calF_\ell,\calM_\ell$ for $\ell=1,\hdots,L$ with coarsest level $L$. We start on finest level $\ell=1$ and subsequently halve image size and resolution from level to level. Registration starts on coarsest level $L$ and we compute deformation $y_L$ from images $\calF_L$ and $\calM_L$ as network input. On all finer levels $\ell<L$, we incorporate the deformations from all preceding coarse levels as initial guess. Therefore, we combine them by functional composition and warp the moving image at current level. Let $X_\ell$ denote the cell-centered image grid on level $\ell$, we compute the warped moving $\calM_\ell(Y_\ell)$ with 
\begin{equation*}
    Y_\ell := y_{\ell+1} \circ y_{\ell+2} \circ \hdots \circ \y_L(X_\ell)
\end{equation*}
and use it together with fixed image $\calF_\ell$ as network input, yielding the deformation field $y_\ell$ on the current level. The final output deformation $\y$ is then given by composition of the whole sequence of coarse-to-fine solutions, i.e., $\y=\y_1\circ\y_2\circ\hdots\circ\y_L$. To evaluate deformations and images at non-grid grid points, we use trilinear interpolation. Our scheme is summarized in Algorithm~\ref{alg:multilevel}. 

In our experiments we use in particular a three level scheme ($L=3$). and we create image pyramid with three reduced resolution images generated from the original 3D images by applying a low-pass filter with a stride of two, four and eight. During training, the three networks are learned progressively. First, the network on the coarsest level is trained for a fixed amount of epochs. Afterwards, the parameters of the middle network are learned while the coarsest network stays fixed and is only used to produce the initial deformation field. The same procedure is repeated on the finest level. The same architecture is used on all levels. The convolution parameters on the coarsest level are initialized with Xavier uniform \cite{xavier2010initialization}. Whereas, all other networks are using the learned parameters of the previous network as initialization. Note that the receptive field in voxel is the same for all used networks, however, due to to the decreased resolution on the coarse levels, the receptive field in mm is much higher.
\begin{algorithm}[t]\scriptsize
\SetKwInOut{Input}{IN}
\SetKwInOut{Output}{OUT}
 
\Input{Fixed image $\calF$, moving image $\calM$, image grid $X$}

\Output{Corse-to-fine deformations $y_L, ..., y_1$, transformed grid $Y=\y_1\circ\hdots\circ\y_L(X)$}

\BlankLine

Create image pyramid $\calF_\ell, \calM_\ell$ for $\ell=1,2,...,L$ with finest level $\ell=1$ and $L$ coarsest.
 
On coarsest level Compute deformation $y_L=\text{CNN}(\calF_L,\calM_L)$

\For{$\ell=L-1,L-2,...,1$}{
    Compute transformed grid $Y_{\ell}= y_{\ell+1} \circ ... \circ y_L(X_\ell)$
    
    Compute deformation $y_\ell = \text{CNN}(\calF_\ell,\calM_\ell(Y_\ell)) $
}
\caption{Multilevel Deep Learning Registration}
\label{alg:multilevel}
\end{algorithm}
\begin{figure}
\centering
\includegraphics[width=0.9\textwidth]{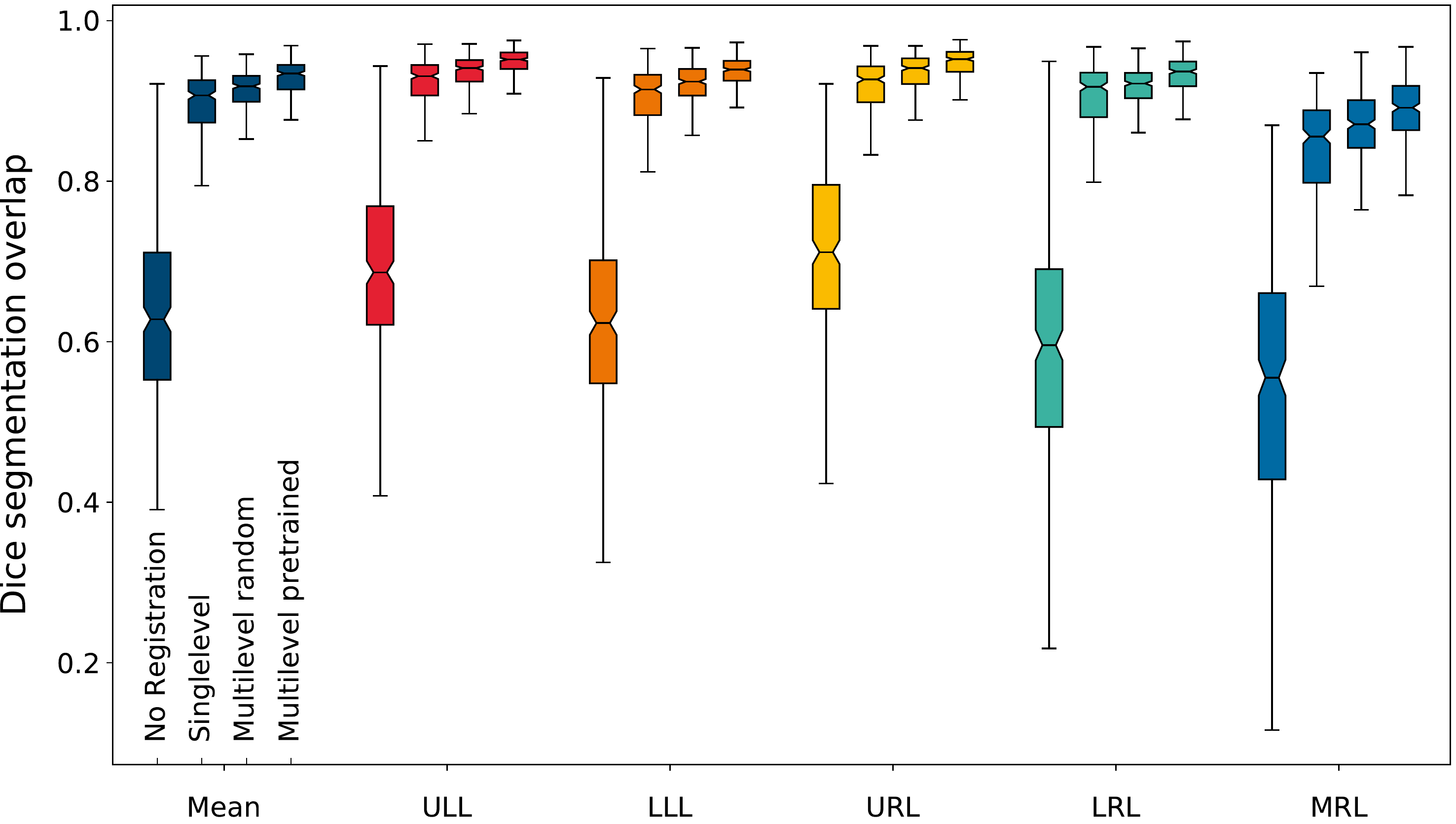}
\caption{Comparison of Dice overlaps for all test images and each anatomical label (average of all labels~\textcolor{col_mean}{\rule[-0.3mm]{.3cm}{.3cm}}, upper left lobe (ULL)~\textcolor{col1}{\rule[-0.3mm]{.3cm}{.3cm}}, lower left lobe (LLL)~\textcolor{col2}{\rule[-0.3mm]{.3cm}{.3cm}}, upper right lobe (URL)~\textcolor{col3}{\rule[-0.3mm]{.3cm}{.3cm}}, lower right lobe (LRL)~\textcolor{col4}{\rule[-0.3mm]{.3cm}{.3cm}}, middle right lobe (MRL) (LRL)~\textcolor{col5}{\rule[-0.3mm]{.3cm}{.3cm}}). For each one the distributions of Dice coefficients after an before registration, after single level dl registration, multilevel dl registration without pretrained CNNs and after multilevel registration with pretrained CNNs.}
\label{fig:Boxplot}
\end{figure}
\section{Experiments and Results}
We demonstrate our deep learning based registration method by registration of inhale-to-exhale  lung CT scans. We use data from 500 patients for training and a disjoint set of 50 patients for validation from the COPDGene study, a large multi-center clinical trial with over 10,000 subjects with chronic obstructive pulmonary disease (COPD) \cite{regan2011COPDGene}. The dataset was acquired across 21 imaging centers using a variety of scanner makes and models. Each patient had received two breath-hold 3D CT scans, one on full inspiration (200mAs) and one at the end of normal expiration (50mAs). For all scans segmentations of the lobes are available, which were computed automatically and manually corrected and verified by trained human analysts. The original images have sizes in the range of $512\times 512 \times \{ 430,\hdots, 901\}$ voxels. Due to memory and time limitations, we create low-resolution images by resampling to a fixed size of $160\times160\times160$ voxels. The low-resolution images are then used during training for the computation of the deformation field and for evaluating of the loss function. Note that, our method is generally not limited to any fixed input size. Although we use  images with $160^3$ voxels, the computed deformation field are defined on full field-of-view and can be evaluated on grids with arbitrary resolution by using trilinear interpolation. Consequently, we use original  full-resolution images for the evaluation of our method.

\subsection*{Multilevel vs. Single Level} 
First, we evaluate our multilevel approach on a disjoint subset of 270 patients from the COPDGene study. We compare our proposed method against a single level approach with only one U-Net using the images on finest level as inputs. We train both approaches for 75 epochs with the same hyper-parameters. For the multilevel approach, the epochs are split equally at each level. We also evaluate the effect on how the network parameter are initialized. Therefore, we compare a Xavier initialization for all convolution parameters of all three networks against our proposed progressive learning strategy. Therefore, only the convolution parameters on the coarsest level are initialized with Xavier initialization and the training of subsequent network is started with the learned parameters form the network of the previous level. 

\begin{figure}[t]
\centering
\setlength{\tabcolsep}{0.001\textwidth}
\begin{tabular}{ccccccc}
  \includegraphics[width=0.165\textwidth]{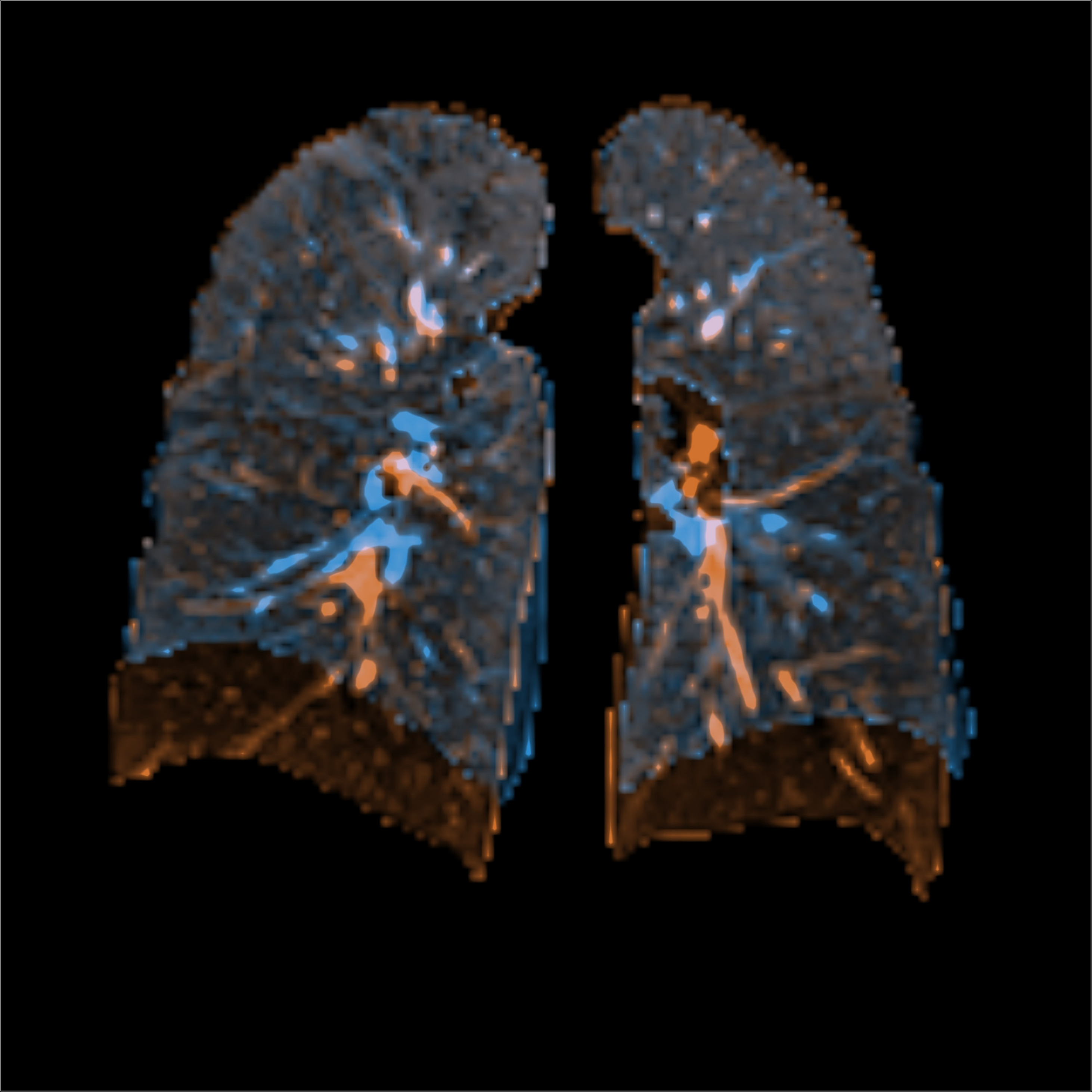}
& \includegraphics[width=0.165\textwidth]{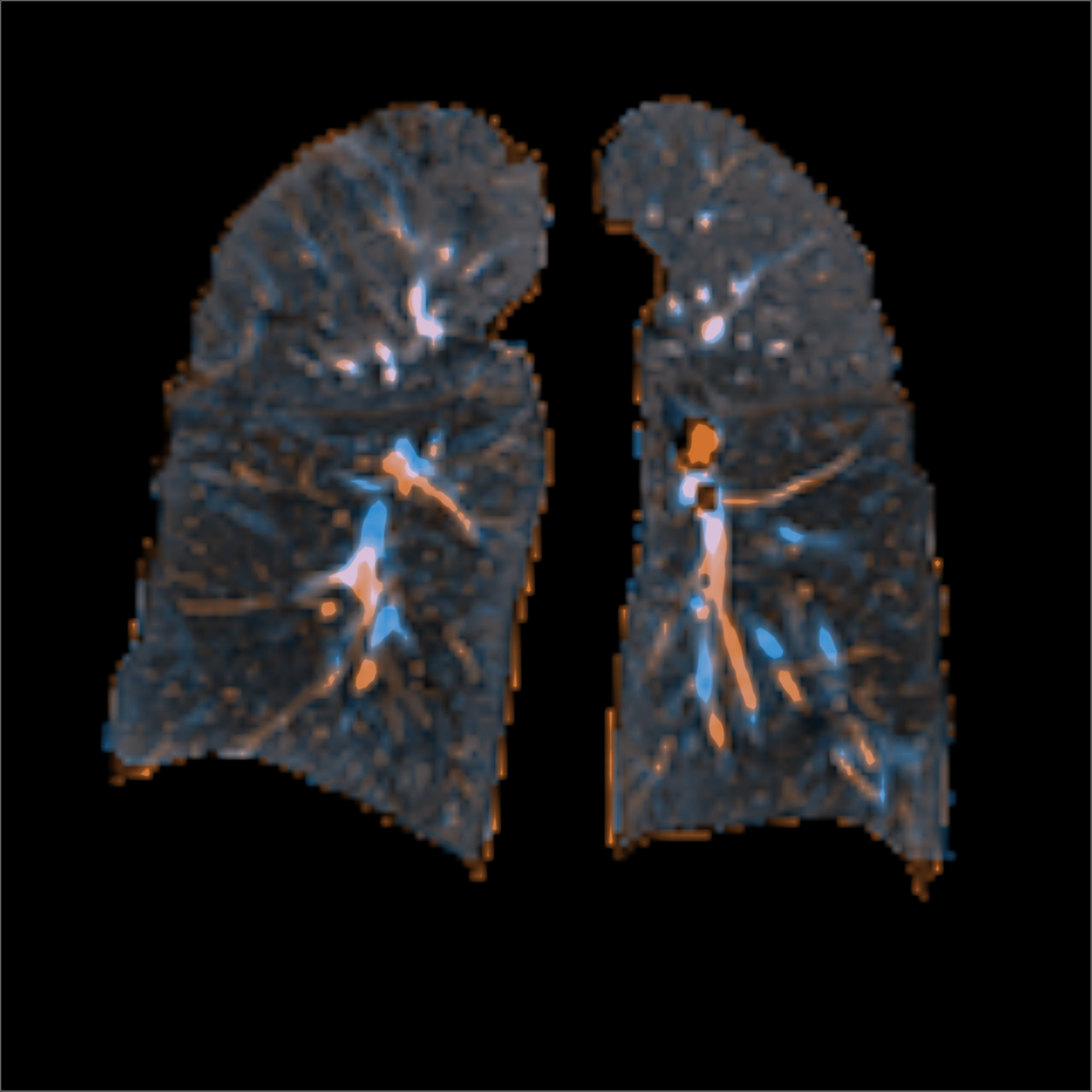}
& \includegraphics[width=0.165\textwidth]{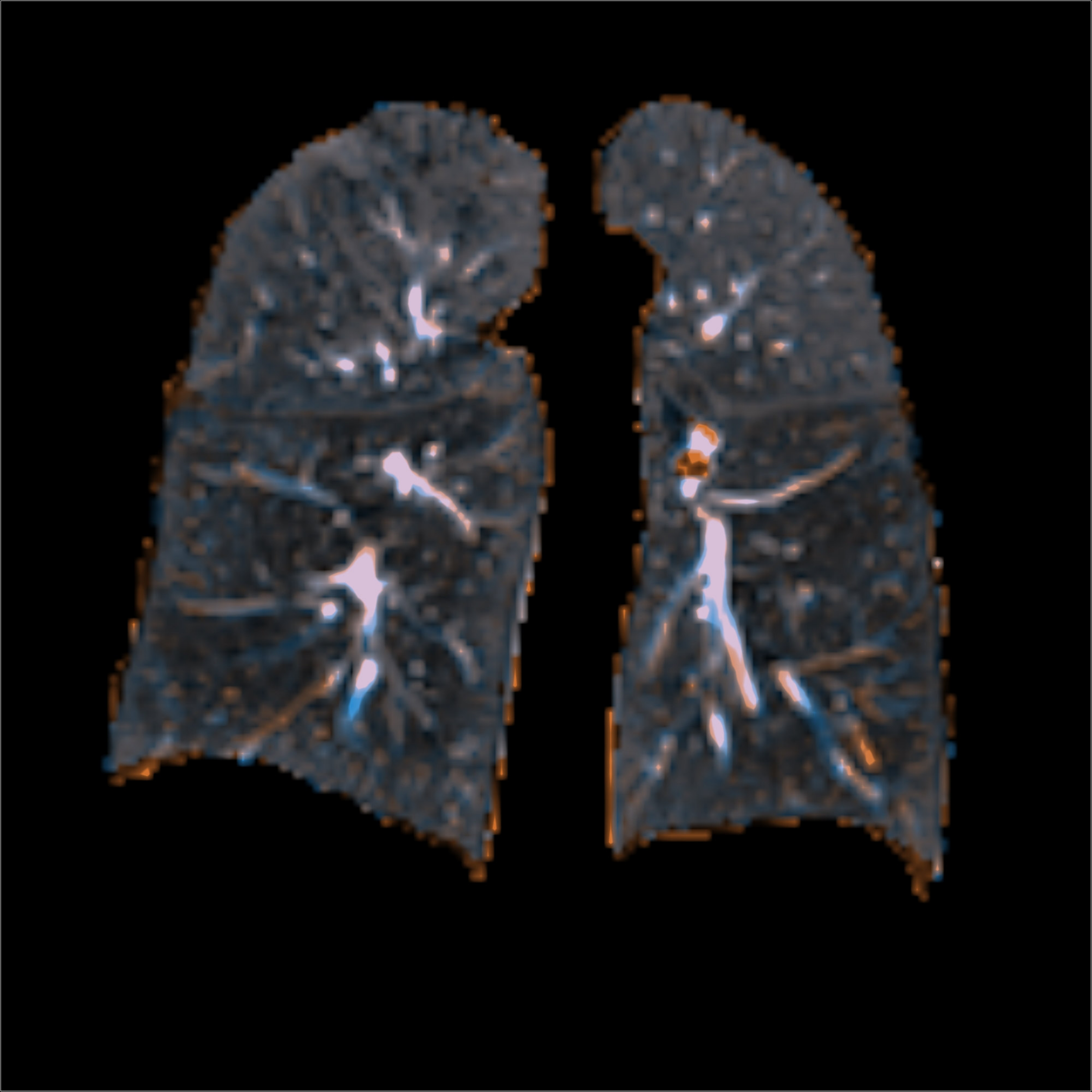}
& \includegraphics[width=0.165\textwidth]{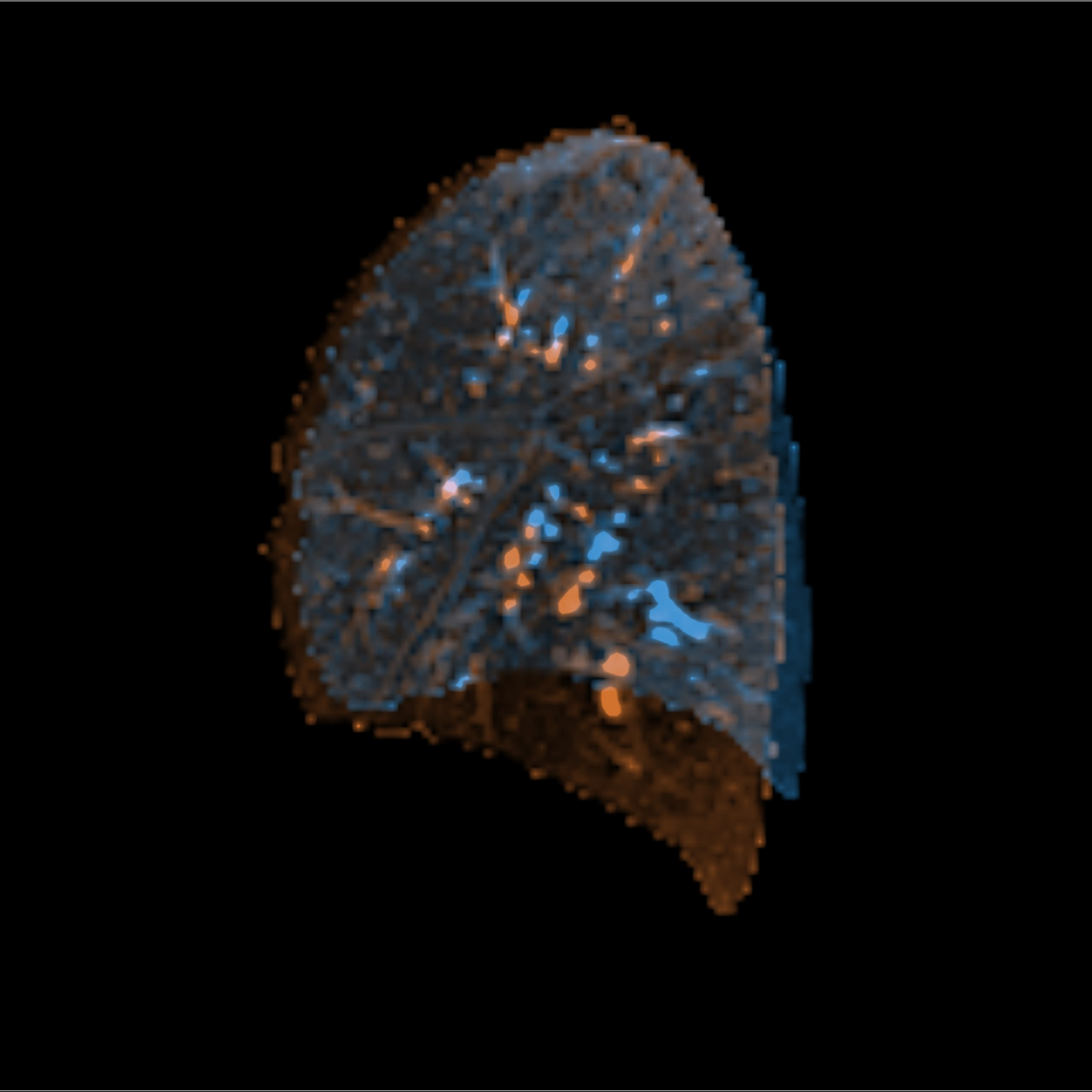}
& \includegraphics[width=0.165\textwidth]{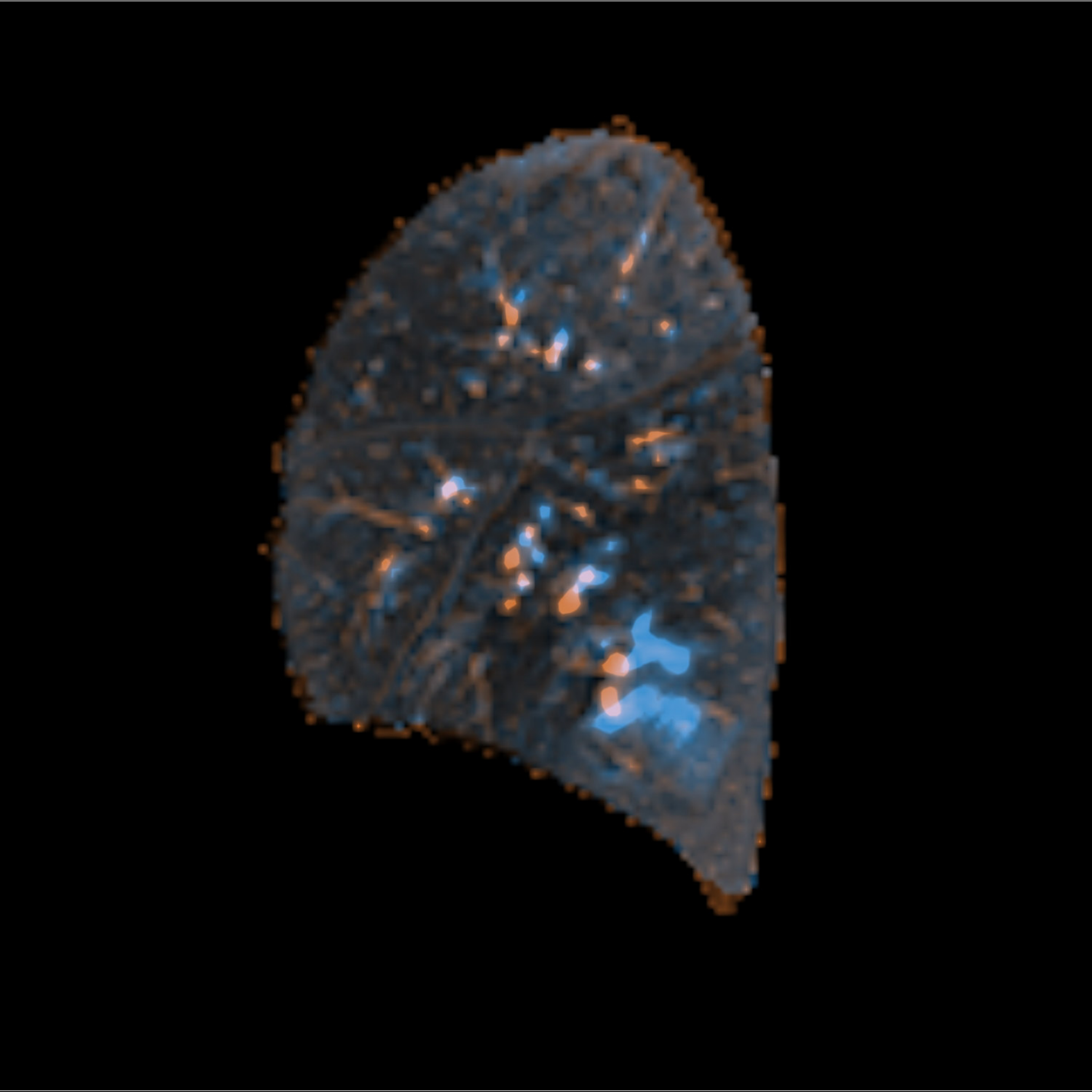}
& \includegraphics[width=0.165\textwidth]{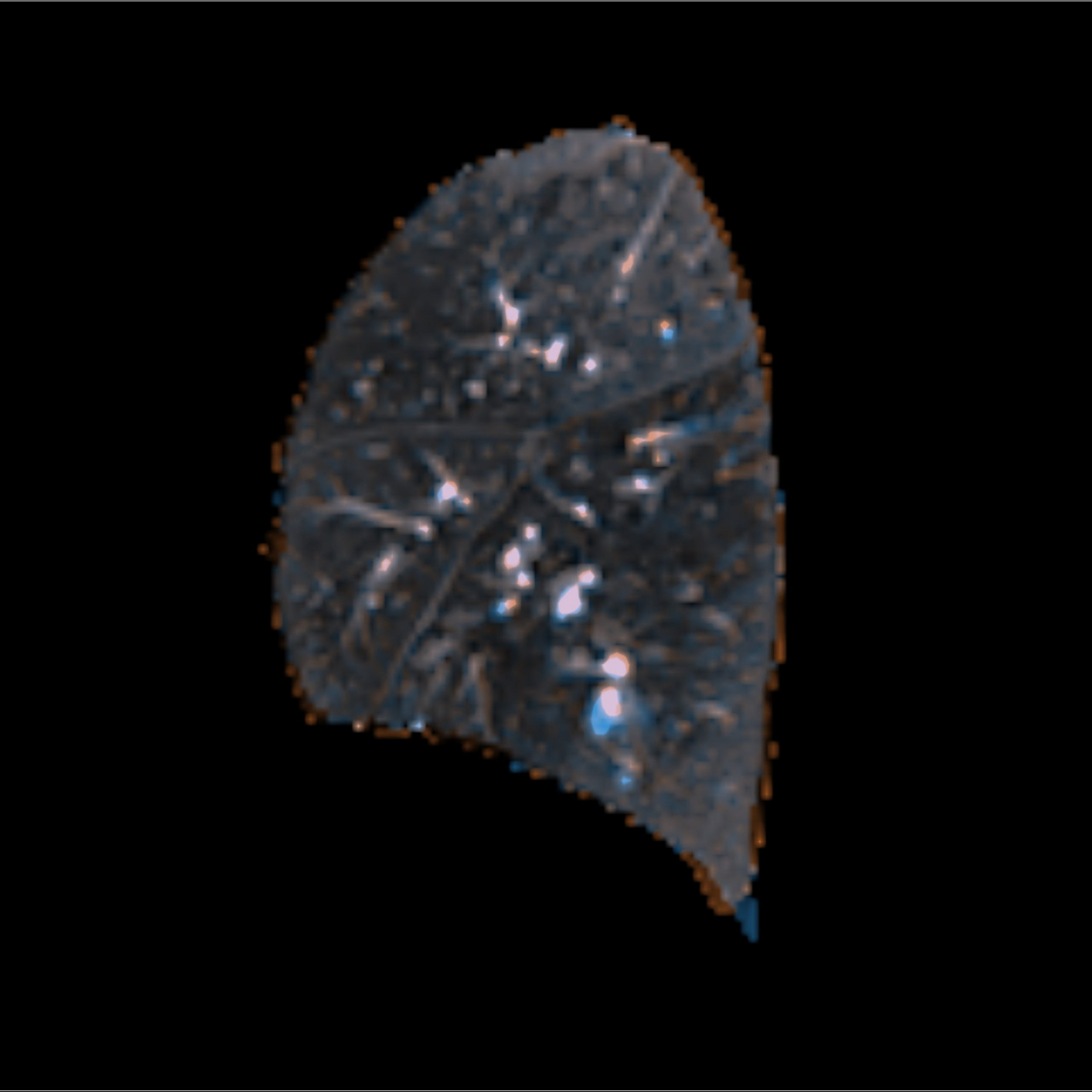}
\\
  \includegraphics[width=0.165\textwidth]{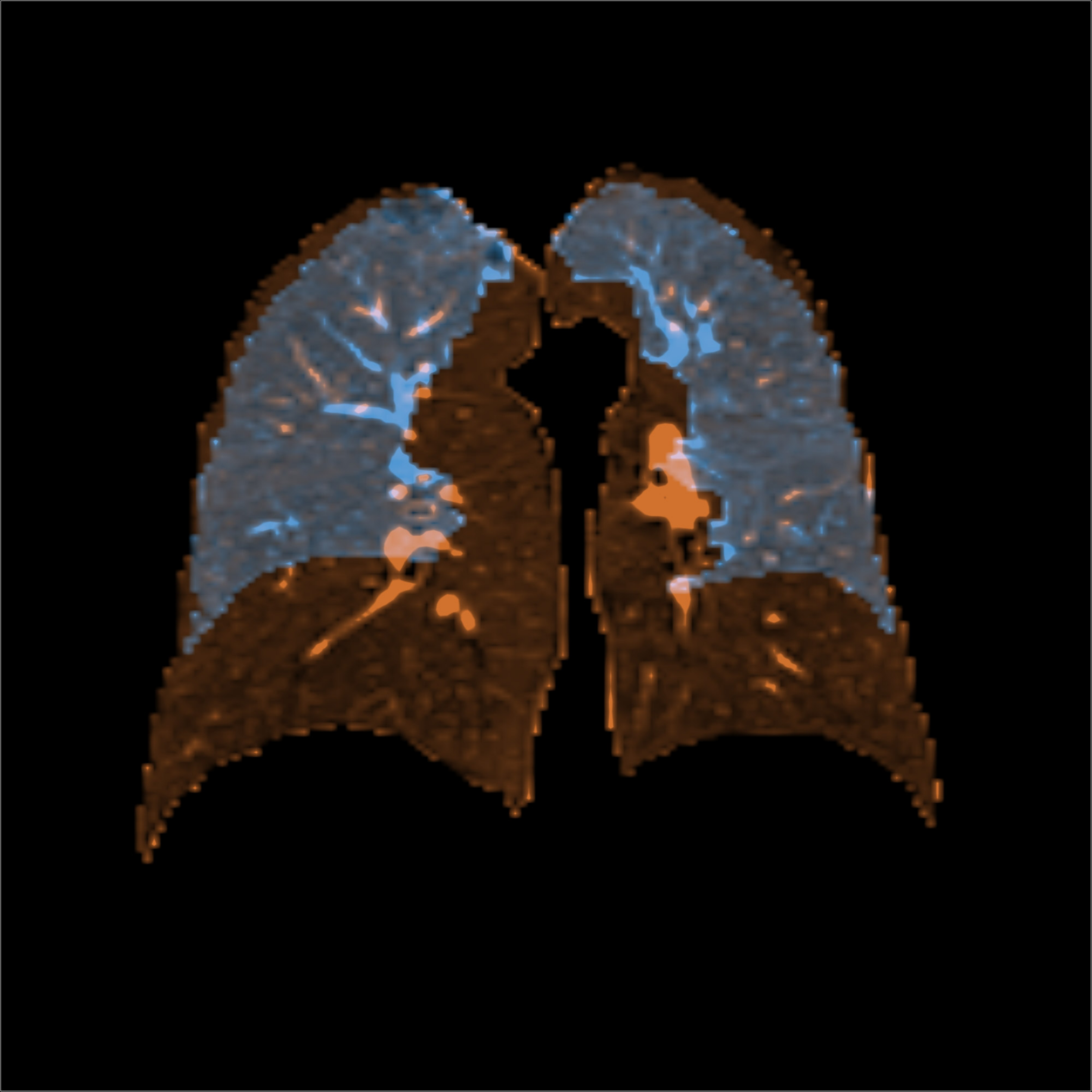}
& \includegraphics[width=0.165\textwidth]{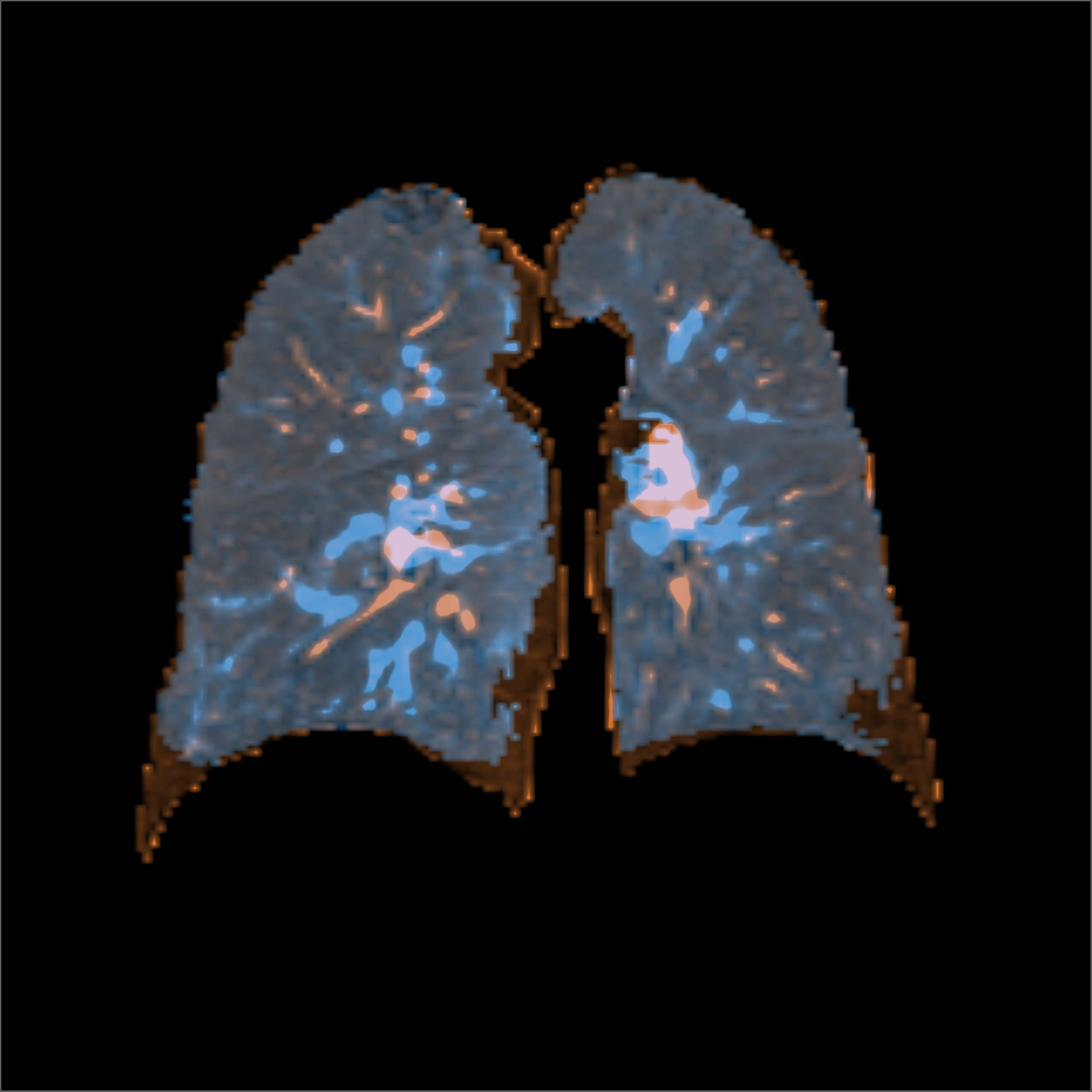}
& \includegraphics[width=0.165\textwidth]{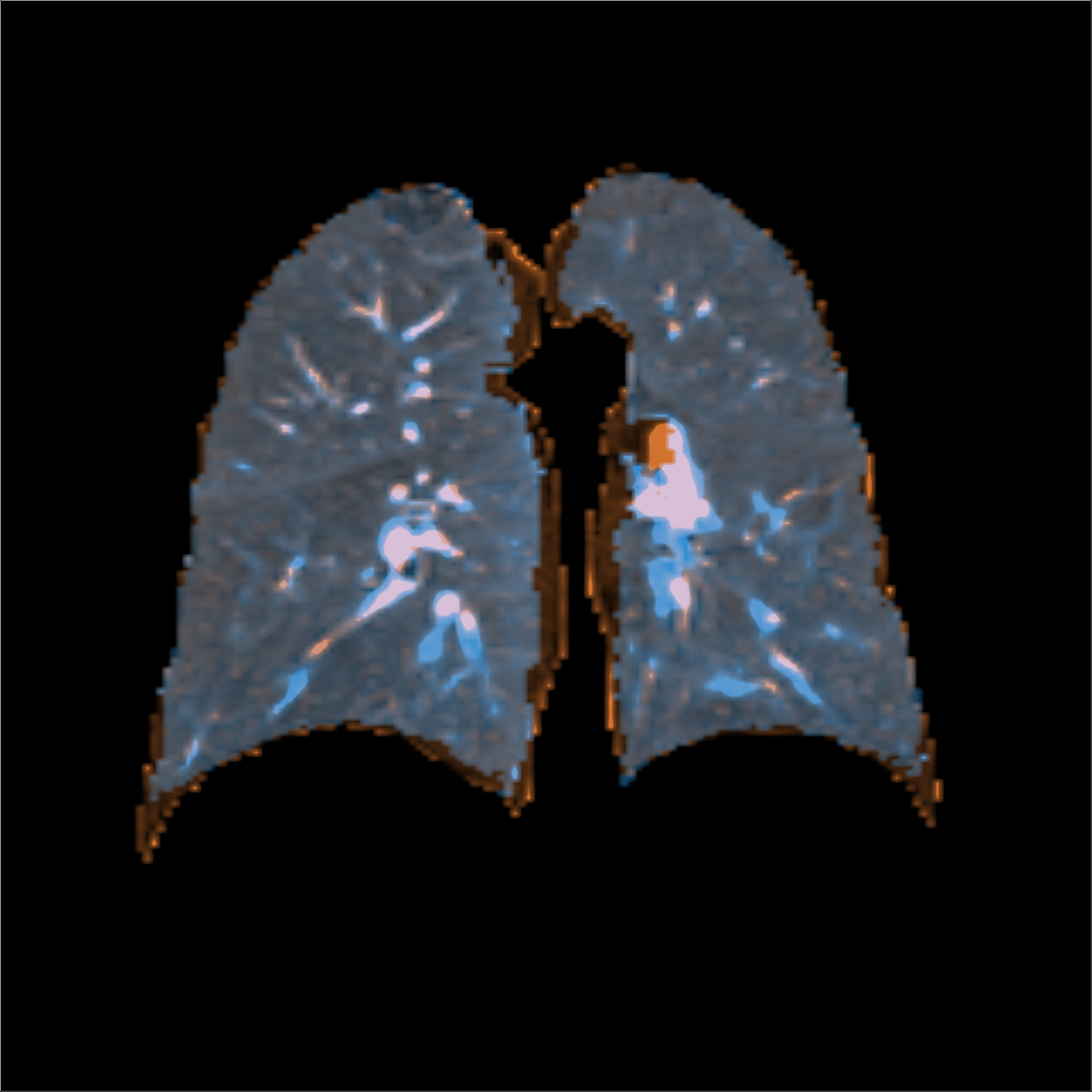}
& \includegraphics[width=0.165\textwidth]{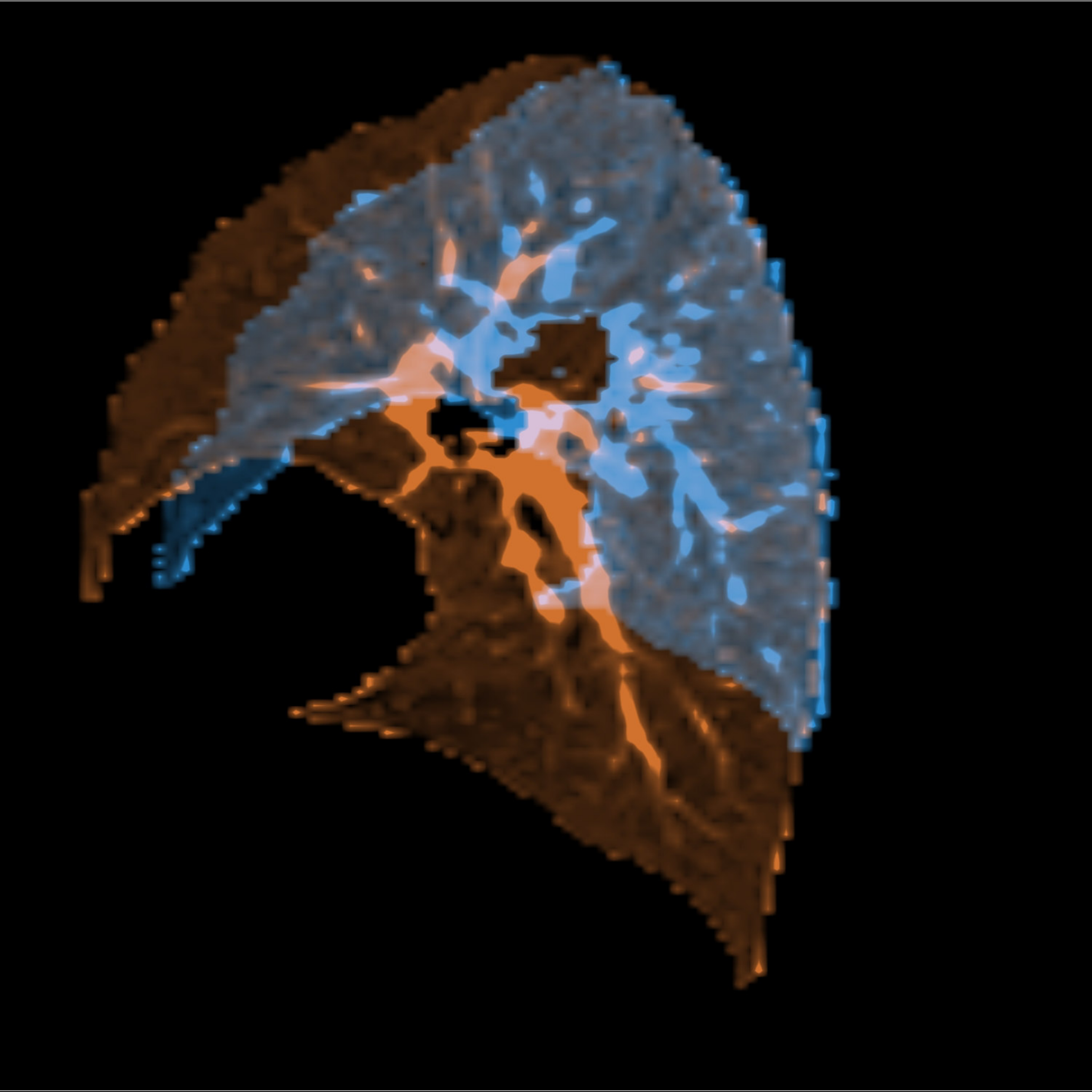}
& \includegraphics[width=0.165\textwidth]{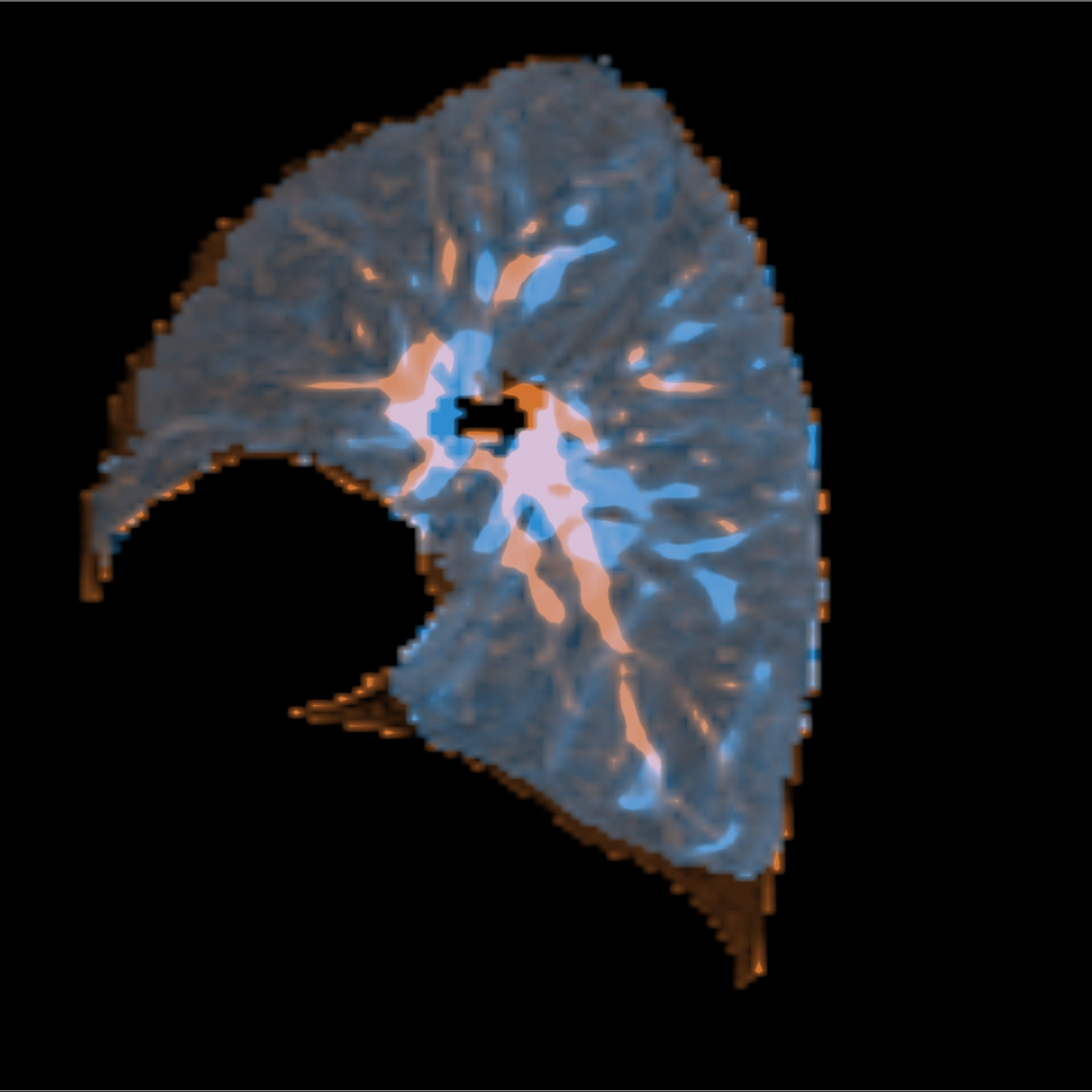}
& \includegraphics[width=0.165\textwidth]{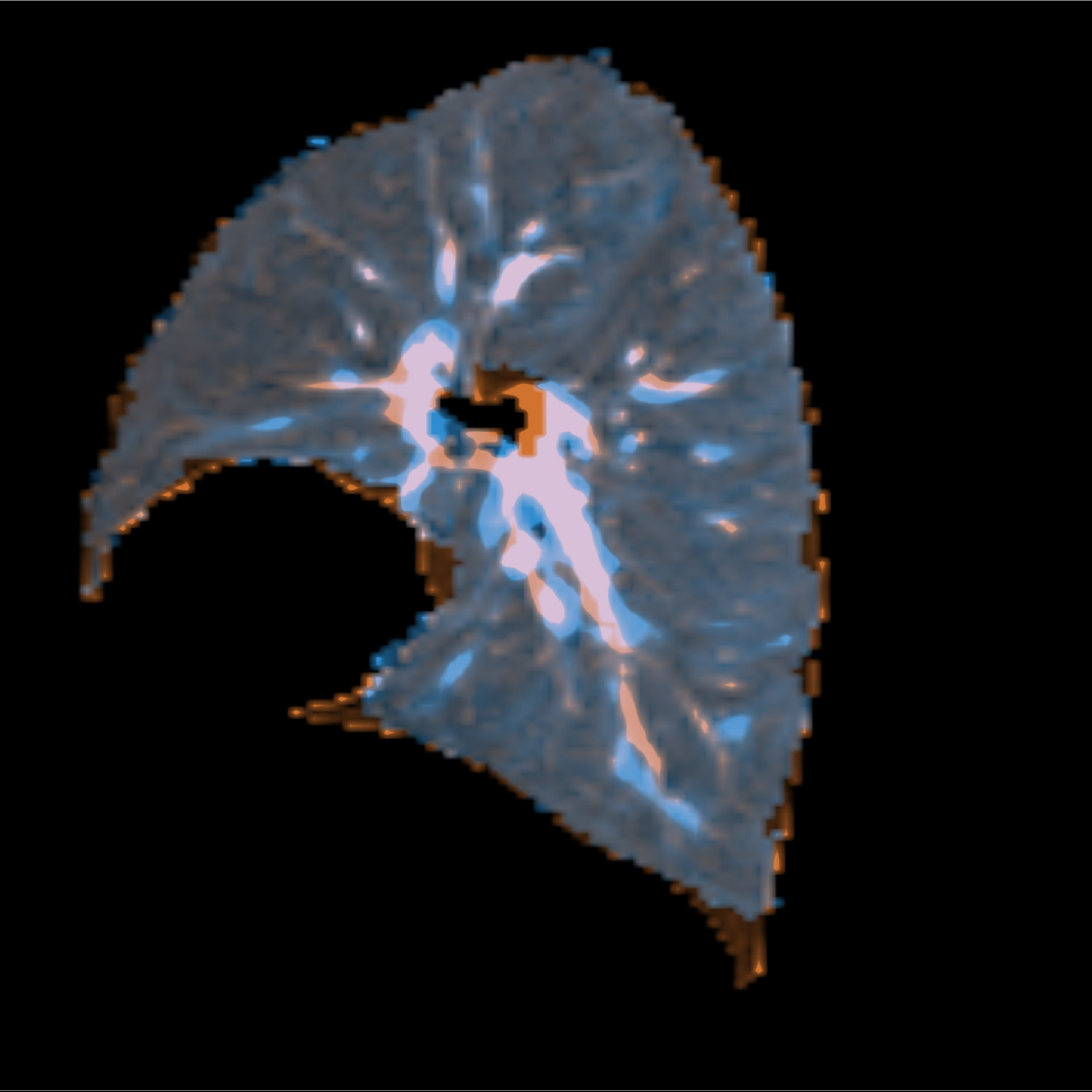}
\\
  {(a) before}
& {(b) single level}
& {(c) multilevel }
& {(d) before }
& {(e) single level}
& {(f) multilevel}
\end{tabular}
\caption{Visualization of two inspiration-expiration registration results: (a) - (c) show coronal views before and after single and multilevel registration, (d) and (f) sagittal views, respectively. The color overlays show the inhale scan in orange and the exhale in blue; due to addition of RGB values, aligned structures appear gray or white. In both cases the respiratory motion was successfully recovered. However, the single level registration does not well align the inner structures.} \label{fig:results}
\end{figure}

We evaluate our method by measuring the overlap of the lobe masks. The underlying assumption is, that if a deformation yields accurate spatial correspondences, then lobe segmentations of the fixed and the warped lobe segmentation of the moving image should overlap well. Fig.~\ref{fig:Boxplot} shows the Dice scores for each label and the average over all labels as a box-plot. Our proposed multilevel approach increase the Dice Score from $63.5\%$ to $92.1\%$. In contrast, the single level method archive a Dice Score of $88.3\%$. Furthermore, the multilevel approach produced less foldings ($0.3\%$ to 2.1\% ). Figure  \ref{fig:results} shows representative qualitative results for of two scan pairs before registration and after our single level and multilevel registration. In both cases the respiratory motion was successfully recovered. 
Although the single level registration produces reasonable Dice scores, it does not well align the inner structures. This is also reflected by the landmark errors in the following section. Comparing the results of the pretrained initialization to the random initialization, an improvement of about $2\%$ in terms of the Dice Score could be reached. 
%
%
%

\subsection*{Comparison with state-of-the-art} 
Additionally, we evaluate our method and compare it to others on the public available DIR-LAB dataset \cite{castillo2013DIRLAB}. It is a collection of ten inspiration-expiration cases with 300 expert-annotated landmarks in the lung. The landmarks are used for evaluating our deformable registration method. The mean (and standard deviation) for all ten scans for the deep learning based multi-resolution approaches of Eppenhof \cite{eppenhof2019progressively} and de Vos (DLIR) \cite{vos2019DLIR}, the single VIRNET and our proposed method are listed in table~\ref{tab:DIRLAB_landmarks}. The overall average landmark error is $2.19\text{mm}$ with a standard deviation of $1.62\text{mm}$. In contrast to the other methods, our mlVIRNET is more robust against outliers and can better handle large initial landmark distances without training on this specific dataset.
\begin{table}[t]
\begin{tabular*}{\linewidth}{l @{\extracolsep{\fill}} ccccc}
\toprule 
Scan & Initial & Eppenhof \cite{eppenhof2019progressively} & DLIR \cite{vos2019DLIR}  & single VIRNET & mlVIRNET \\
\midrule
Case 1 & 3.89(2.78) & 2.18(1.05)  & 1.27(1.16) & 1.73(0.83) &1.33(0.73)\\
Case 2 & 4.34(3.90) &2.06(0.96) & 1.20(1.12) &2.38(1.11) &1.33(0.69)\\
Case 3 & 6.94(4.05) &2.11(1.04) &1.48(1.26) &3.01(1.86) &1.48(0.94)\\
Case 4 & 9.83(4.85) &3.13(1.60) & 2.09(1.93)&4.28(2.37) &1.85(1.37) \\
Case 5 & 7.48(5.50)& 2.92(1.70)& 1.95(2.10)& 3.17(2.2)&1.84(1.39)\\
Case 6 & 10.89(6.96)&4.20(2.00) & 5.16(7.09)&4.85(3.04) & 3.57(2.15) \\
Case 7 & 11.03(7.42)&4.12(2.97) & 3.05(3.01)&3.67(1.82) &2.61(1.63)\\
Case 8 &14.99(9.00)& 9.43(6.28)& 6.48(5.37)& 5.75(3.93) &2.62(1.52)\\
Case 9 &7.92(3.97) & 3.82(1.69)& 2.10(1.66)& 4.90(2.25)&2.70(1.46)\\
Case 10 & 7.30(6.34)&2.87(1.96) & 2.09(2.24)&3.49(2.21) &2.63(1.93)  \\\bottomrule
Total & 8.46(6.58) & 3.68(3.32)& 2.64(4.32)& 3.72(2.45)&2.19(1.62)\\
\end{tabular*}
\normalsize
\caption{Mean (standard deviation) of the registration error in mm determined on DIR-Lab 4D-CT data. From left to right: initial landmark error, the multi-resolution approaches of \cite{eppenhof2019progressively} and \cite{vos2019DLIR} and the single level VIRNET and the proposed multilevel VIRNET.}
\label{tab:DIRLAB_landmarks}
\end{table}

\section{Discussion and Conclusion}
We presented an end-to-end multilevel framework for deep learning based image registration which is able to compensate and handle large deformations by computing deformation fields on different scales. Our method takes the whole image information into account and predicts a dense 3D deformation field. We validated our framework on the challenging task of large motion inhale-to-exhale registration using large image data of the multi-center COPDGene study. We have shown that our proposed method archives better results than the comparable single level variant. In particular with regard to the alignment of inner lung structures and the presence of foldings. Only less than $0.3\%$ voxel positions of the images showed a folding.  Additionally, we demonstrated that using the network parameter of the previous level as initialization, yields to better registration results. Moreover, we demonstrated the transferability of our approach to new datasets by evaluating our learned method on the publicly available DIRLAB dataset and showing a lower landmark error than other deep learning based registration methods.\\
\ \\
\textbf{Acknowledgements:} 
We  gratefully  acknowledge  the  COPDGene  Study  for providing the data used.  COPDGene is funded by Award
Number U01 HL089897 and Award Number U01 HL089856 from the National Heart, Lung, and Blood Institute. The COPDGene  project  is  also  supported  by  the  COPD  Foundation through contributions made to an Industry Advisory Board comprised of AstraZeneca, Boehringer Ingelheim, GlaxoSmithKline, Novartis, Pfizer, Siemens and Sunovion.
%
%
%
\bibliographystyle{splncs04}
\bibliography{paper2192}

\begin{thebibliography}{10}
\providecommand{\url}[1]{\texttt{#1}}
\providecommand{\urlprefix}{URL }
\providecommand{\doi}[1]{https://doi.org/#1}

\bibitem{bajcsy1989multiresolution}
Bajcsy, R., Kova{\v{c}}i{\v{c}}, S.: Multiresolution elastic matching. Computer
  vision, graphics, and image processing  \textbf{46}(1),  1--21 (1989)

\bibitem{balakrishnan2019voxelmorph}
Balakrishnan, G., Zhao, A., Sabuncu, M.R., Guttag, J., Dalca, A.V.: Voxelmorph:
  a learning framework for deformable medical image registration. IEEE TMI
  (2019)

\bibitem{castillo2013DIRLAB}
Castillo, R., Castillo, E., Fuentes, D., Ahmad, M., Wood, A.M., Ludwig, M.S.,
  Guerrero, T.: A reference dataset for deformable image registration spatial
  accuracy evaluation using the copdgene study archive. Physics in Medicine \&
  Biology  \textbf{58}(9), ~2861 (2013)

\bibitem{eppenhof2019progressively}
Eppenhof, K.A., Lafarge, M.W., Pluim, J.P.: Progressively growing convolutional
  networks for end-to-end deformable image registration. In: Medical Imaging
  2019: Image Processing. vol. 10949, p. 109491C. International Society for
  Optics and Photonics (2019)

\bibitem{xavier2010initialization}
Glorot, X., Bengio, Y.: Understanding the difficulty of training deep
  feedforward neural networks. In: Proceedings of the thirteenth international
  conference on artificial intelligence and statistics. pp. 249--256 (2010)

\bibitem{hering2019unsupervised}
Hering, A., Heldmann, S.: Unsupervised learning for large motion thoracic ct
  follow-up registration. In: SPIE Medical Imaging: Image Processing. vol.
  10949, p. 109491B (2019)

\bibitem{HeringEtAl2019}
Hering, A., Kuckerts, S., Heldmann, S., Heinrich, M.P.: Enhancing label-driven
  deep deformable image registration with local distance metrics for
  state-of-the-art cardiac motion tracking. In: BVM 2019. pp. 309--314.
  Springer (2019)

\bibitem{hu2018labeldriven}
Hu, Y., Modat, M., Gibson, E., Ghavami, N., Bonmati, E., Moore, C.M., Emberton,
  M., Noble, J.A., Barratt, D.C., Vercauteren, T.: Label-driven
  weakly-supervised learning for multimodal deformable image registration. In:
  Proc. of ISBI 2018. pp. 1070--1074. IEEE (2018)

\bibitem{kabus2010multilevel}
Kabus, S., Lorenz, C.: Fast elastic image registration. Medical Image Analysis
  for the Clinic: A Grand Challenge pp. 81--89 (2010)

\bibitem{Modersitzki2009}
Modersitzki, J.: FAIR: Flexible Algorithms for Image Registration. SIAM (2009)

\bibitem{haber2004cofir}
Modersitzki, J., Haber, E.: COFIR: Coarse and Fine Image Registration,
  chap.~14, pp. 277--288. Computational Science \& Engineering: Real-Time
  PDE-Constrained Optimization, SIAM (2007)

\bibitem{regan2011COPDGene}
Regan, E.A., Hokanson, J.E., Murphy, J.R., Make, B., Lynch, D.A., Beaty, T.H.,
  Curran-Everett, D., Silverman, E.K., Crapo, J.D.: Genetic epidemiology of
  copd (copdgene) study design. COPD: Journal of Chronic Obstructive Pulmonary
  Disease  \textbf{7}(1),  32--43 (2011)

\bibitem{rohe2017svf}
Roh{\'e}, M.M., Datar, M., Heimann, T., Sermesant, M., Pennec, X.: Svf-net:
  Learning deformable image registration using shape matching. In: MICCAI 2017.
  pp. 266--274. Springer (2017)

\bibitem{ruhaak2017estimation}
R{\"u}haak, J., Polzin, T., Heldmann, S., Simpson, I.J., Handels, H.,
  Modersitzki, J., Heinrich, M.P.: Estimation of large motion in lung ct by
  integrating regularized keypoint correspondences into dense deformable
  registration. IEEE TMI  \textbf{36}(8),  1746--1757 (2017)

\bibitem{schnabel2001generic}
Schnabel, J.A., Rueckert, D., Quist, M., Blackall, J.M., Castellano-Smith,
  A.D., Hartkens, T., Penney, G.P., Hall, W.A., Liu, H., Truwit, C.L., et~al.:
  A generic framework for non-rigid registration based on non-uniform
  multi-level free-form deformations. In: MICCAI 2001. pp. 573--581. Springer
  (2001)

\bibitem{vos2019DLIR}
de~Vos, B.D., Berendsen, F.F., Viergever, M.A., Sokooti, H., Staring, M.,
  I{\v{s}}gum, I.: A deep learning framework for unsupervised affine and
  deformable image registration. Medical image analysis  \textbf{52},  128--143
  (2019)

\end{thebibliography}

\end{document}